\pdfoutput=1
\documentclass{article}

\PassOptionsToPackage{numbers, compress}{natbib}

\usepackage[preprint]{neurips_2023}

\usepackage[utf8]{inputenc} 
\usepackage[T1]{fontenc}    
\usepackage{hyperref}       
\usepackage{url}            
\usepackage{booktabs}       
\usepackage{amsfonts}       
\usepackage{nicefrac}       
\usepackage{microtype}      
\usepackage{xcolor}         

\usepackage{amsmath}
\usepackage{multirow}
\usepackage{tikz}
\usepackage{color}
\usepackage{pifont}
\usepackage{amssymb}
\usepackage{enumitem}
\usepackage{xspace}

\usepackage{graphicx}
\usepackage{caption} 
\usepackage{subcaption}
\usepackage{wrapfig}
\usepackage[ruled,vlined]{algorithm2e}
\usepackage{algpseudocode}

\usepackage{graphicx}
\usepackage{amsmath}
\usepackage{amssymb}
\usepackage{booktabs}
\usepackage{adjustbox}
\usepackage{multirow}
\usepackage{arydshln}
\usepackage{colortbl}
\usepackage{amsfonts}       
\usepackage{nicefrac}       
\usepackage{microtype}      
\usepackage{times}
\usepackage{epsfig}
\usepackage{tipa}
\usepackage[T1, OT1]{fontenc}
\DeclareTextSymbolDefault{\dh}{T1}
\usepackage{booktabs}
\usepackage{multirow}
\usepackage{color, colortbl}
\usepackage{pifont}
\usepackage{makecell}
\usepackage{threeparttable}
\usepackage{xcolor}
\usepackage{scalerel}
\usepackage{makecell}
\usepackage{tabu}
\usepackage{tikz}
\usepackage{marvosym}
\usepackage{xcolor} 

\definecolor{Gray}{gray}{0.5}
\definecolor{LGray}{gray}{0.9}
\definecolor{darkblue}{RGB}{94,110,186}
\definecolor{darkGreen}{RGB}{92, 148, 110}
\definecolor{myblue}{RGB}{14, 121, 178}
\definecolor{rightpath}{RGB}{248, 203, 173}
\definecolor{wrongpath}{RGB}{89, 89, 89}

\newcommand{\darkGreen}[1]{\textcolor{darkGreen}{#1}}

\usepackage{tabulary}
\usepackage{tabulary,multirow,overpic,xcolor}
\newcolumntype{x}[1]{>{\centering\arraybackslash}m{#1pt}}
\newcolumntype{y}[1]{>{\arraybackslash}m{#1pt}}

\title{VideoChat-R1: Enhancing Spatio-Temporal Perception via Reinforcement Fine-Tuning}

\author{Xinhao Li$^{2,1*}$, Ziang Yan$^{3,1*}$, Desen Meng$^{2*}$, Lu Dong$^{4,1}$, Xiangyu Zeng$^{2,1}$, Yinan He$^{1}$ \\ \textbf{Yali Wang}$^{6,1}$, \textbf{Yu Qiao}$^{1,6}$, \textbf{Yi Wang}$^{\dagger1,5}$, \textbf{Limin Wang}$^{\dagger2,1}$  \\
    \small$^1$Shanghai AI Laboratory~~~
    \small$^2$Nanjing University~~~ 
    \small$^3$Zhejiang University \\
    \small$^4$University of Science and Technology of China ~~~\small$^5$Shanghai Innovation Institute\\
    \small$^6$Shenzhen Institutes of Advanced Technology, Chinese Academy of Sciences \\
    {\small \url{https://github.com/OpenGVLab/VideoChat-R1}}
}

\begin{document}

\maketitle
{
\renewcommand{\thefootnote}%
{\fnsymbol{footnote}}
\footnotetext[0]{* Equal contribution. $\dagger$ Corresponding authors.} 
}

\begin{abstract}
Reinforcement Learning (RL) benefits Large Language Models (LLMs) for complex reasoning. Inspired by this, we explore integrating spatio-temporal specific rewards into Multimodal Large Language Models (MLLMs) to address the unique challenges of video understanding, such as long-range temporal associations. This paper investigates how rule-based rewards, particularly temporal ones, can improve video reasoning and their generalizability.
Our study proposes Reinforcement Fine-Tuning (RFT) as a data-efficient method to enhance video reasoning on specific tasks without sacrificing original capabilities. Through joint RFT on multiple spatio-temporal perception tasks, we developed \textbf{VideoChat-R1}, a powerful Video MLLM. VideoChat-R1 achieves state-of-the-art spatio-temporal perception, demonstrating significant improvements in tasks like temporal grounding (\textbf{+31.8}) and object tracking (\textbf{+31.2}), while also improving general QA benchmarks. The enhanced perception and preserved chat abilities contribute to a more reliable video dialogue system, leading to our ``Temporal Clue-driven Reasoning" inference schema. This work provides a foundation for developing robust, real-world video comprehension agents.
\end{abstract}

\section{Introduction}
\label{sec:Introduction}

The integration of reinforcement learning (RL) has notably propelled the capabilities of large language models (LLMs) forward, particularly in complex reasoning. Studies, such as o-series from OpenAI~\cite{jaech2024openai} and R1 of DeepSeek~\cite{guo2025deepseek}, demonstrated that test-time scaling improves model reasoning and such reasoning is elicit-able via only rule-based rewarding.
Inspired by these advances, community is interested in tuning multimodal large language models (MLLMs) with vision-related verifications in Group Relative Policy Optimization (GRPO) \cite{guo2025deepseek} manner ~\cite{zhou2025r1,segzero,visonr1,deng2025boosting,peng2025lmm,visualrft,r1ov,r1vl,zhan2025visionr1,deng2025openvlthinker}. They primarily consider visual mathematical reasoning and spatial localization.

Regarding video understanding, reasoning is also crucial as some of its core abilties (e.g. its long-range temporal association, fine-grained spatiotemporal understanding based on user queries) are barely addressed via perception or simple analysis. 
Initial works~\cite{r1omni,wang2025timezero,videor1} have validated the superiority of the GRPO algorithm over supervised fine-tuning in some specific video tasks, such as temporal grounding and video question answer. 
Considering advancing the agent development with real-world video comprehension, we need to investigate and integrate spatio-temporal specific rewards or verification mechanisms directly into the learning of current MLLMs in a scalable manner. This could foster a more tightly coupled, near-closed-loop learning environment, enabling MLLMs to master intricate spatio-temporal skills. While our current work may not fully achieve this ambitious goal, it endeavors to lay foundational groundwork for the community by exploring how to embed such mechanisms effectively.
Specifically, we in this paper chart the landscape of how rule-based rewards (especially temporal related ones) work in video understanding, along with systematic evaluations of the its generalizability across diverse video-based reasoning scenarios.

\begin{figure*}[t]
	\centering
	\includegraphics[width=1.0\textwidth]{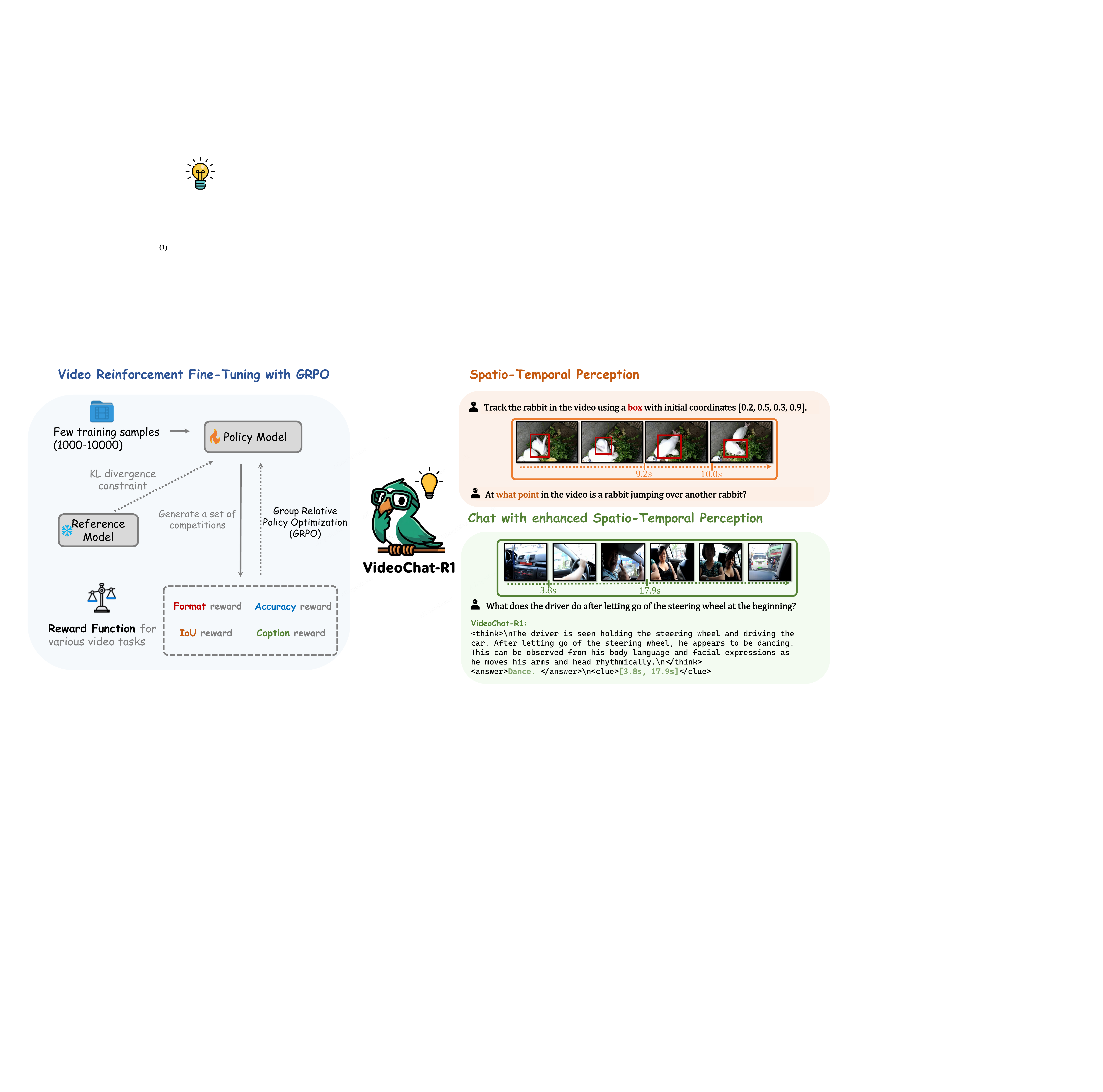} 
	\caption{
    \textbf{Overview of VideoChat-R1.} Through reinforcement learning fine-tuning using GRPO, VideoChat-R1 has powerful spatio-temporal perception capabilities and can apply these capabilities in chatting scenarios. 
    }
	\label{fig1:videor1}
\end{figure*}

Our study begins with spatio-temporal perception, extending to other popular video tasks (e.g. video caption), forming a practical recipe to enhance video reasoning via proper Reinforcement Fine-Tuning (RFT). We evaluate the robustness of various video verifications through different training and testing setting, also benchmarking their respective data requirements. Moreover, how these rewards interact with each other are also systematically explored. Based on these experiences, we give an intuitive and effective inference schema named ``Temporal Clue-driven Reasoning''. Generally, our main findings are as follows.

\begin{itemize}
    \item \textbf{\textit{Reinforcement fine-tuning is data-efficient for enhancing models on specific tasks without sacrificing original capabilities}}. With a small amount of data, training via RFT can yield a remarkable improvement in spatio-temporal perception ability, and there is negligible impact on the performance of out-domain tasks and the original general capabilities of the model, which outperforms traditional supervised fine-tuning significantly. 

    \item \textbf{\textit{Through joint reinforcement fine-tuning on multiple spatio-temporal perception tasks}}, we construct \textbf{VideoChat-R1}, a powerful Video MLLM that boasts state-of-the-art spatiotemporal perception capabilities while also taking into account chat abilities. We have also discovered that training on spatio-temporal perception tasks has slightly strengthened the model's spatio-temporal reasoning abilities. Compared with Qwen2.5-VL-7B,  VideoChat-R1 achieves several times the performance improvement in spatiotemporal perception tasks such as temporal grounding (\textbf{+31.8}) and object track (\textbf{+31.2}). At the same time, it also achieves  non-trivial improvements on general QA benchmarks, such as VideoMME (\textbf{+0.9}), MVBench (\textbf{+1.0}), and Perception Test (\textbf{+0.9})

\item \textbf{\textit{The improvement of spatio-temporal perception ability and the preservation of the original chat capability can contribute to a more reliable and efficient video dialogue system}}. Our VideoChat-R1 can provide reference video segments when answering users' questions. Meanwhile, we propose to utilize these video segments for \textbf{``Temporal Clue-driven Reasoning"} to further obtain more accurate answers. Our experimental results reveal the potential of the approach that enhances the model's spatiotemporal perception ability through reinforcement learning for future research in the directions of reliable video dialogue systems and long video understanding.

\end{itemize}

\section{Related work}
\label{sec:Related_work}

\paragraph{Reasoning in MLLMs.}

The field LLMs has witnessed significant breakthroughs in reasoning, exemplified by recent works \cite{jaech2024openai,guo2025deepseek}. These developments~\cite{shao2024deepseekmath,guo2025deepseek,team2025kimi15} have markedly improved LLMs' proficiency in tackling complex, multi-step tasks, such as challenging mathematical problems and coding exercises. Extending these RL techniques to multimodal large language models (MLLMs) has seen numerous efforts~\cite{zhou2025r1,segzero,visonr1,deng2025boosting,peng2025lmm,visualrft,r1ov,r1vl,zhan2025visionr1,deng2025openvlthinker} focused on leveraging verifiable reward mechanisms to boost visual reasoning performance. However, the application of RL-based strategies to spatio-temporal reasoning within the video domain remains comparatively less explored, with a few studies~\cite{wang2025timezero,r1omni,videor1} investigating this area. \cite{wang2025timezero} and ~\cite{r1omni} show GRPO's potential in temporal grounding and sentiment analysis, while, ~\cite{videor1} demonstrates how GRPO can facilitate temporal reasoning.

\paragraph{Spatio-Temporal Understanding with MLLMs.}

Video understanding heavily relies on spatio-temporal perception. Progress in Video Multimodal Large Language Models (video MLLMs)~\cite{li2023videochat,maaz2023video,li2024mvbench,wang2024internvideo2,zhang2024video,wang2025internvideo25,li2024videochatf,videochatonline,qwen25vl} addresses video question answering and captioning in the unfied dialogue format, leveraging LLMs to organize predictions over visual perceptions. ~\cite{yu2024merlin} and ~\cite{zeng2024timesuite} leverage data augmentation for temporal modeling, yet compromising their general performance. ~\cite{yan2024task} improves fine-grained spatio-temporal perception by task preference optimization with solid supervised fine-tuning using a lot of data.

\section{Methodology}
\label{sec:Methodology}

We present how to exploit the spatio-temporal rewards to improve MLLMs in video domains using GRPO. Before that, we revisit the Group Relative Policy Optimization (GRPO)~\cite{shao2024deepseekmath} first.

\subsection{Preliminary}
\label{subsec:Preliminary}

Group Relative Policy Optimization (GRPO)~\cite{shao2024deepseekmath} compares groups of candidates responses directly, independent on critic models. Regarding this, it significantly lowers training resources. Formally,
with an input query $q$, GRPO initiates by generating a set of 
$G$ distinct candidate responses, denoted as $o=\{o_1, \dots,o_G\}$, via policy sampling. Each of these candidates is then assigned a score from a predefined reward function, yielding ${R_1, \dots, R_G}$. To normalize these scores and ascertain the relative quality of each response, GRPO computes their mean and standard deviation:
\begin{equation}
\label{eq:ro}
    A_i=
    \frac{R_i-\mathrm{mean}(\{R_i\}_{i=1}^G)}{\mathrm{std}(\{R_i\}_{i=1}^G)} \text{.}
\end{equation}
Here $A_i$ quantifies the relative quality of the $i$-th candidate response within its group. The objective of GRPO is to encourage the model to prioritize and generate responses that achieve higher relative scores within such groups. Furthermore, the overall training objective incorporates a KL-divergence term $\mathrm{D}_\mathrm{KL}(\cdot|\cdot)$, which serves to constrain the optimized policy $\pi_\theta$ from diverging excessively from the original MLLM parameters $\pi_\mathrm{ref}$.

\subsection{Spatio-Temporal Rewards of Video MLLM in GRPO}

We explore how to use GRPO to enhance the performance of Video MLLM in video-language understanding. We consider the five most common types of video related tasks: temporal grounding, object tracking, video question answering, captioning, and quality assessment in our experiments.

\paragraph{Format reward.} To enable the model to output responses in the format we desire. For example, we expect the model to enclose its thought process with \texttt{<think>...</think>} and the answer with \texttt{<answer>...</answer>}, we designed a format reward $R_{\mathrm{format}}$ for each task. We use regular expression matching to determine whether the model adheres to our specified format:
\begin{equation}
    R_{\mathrm{format}} = \begin{cases}
1, & \text{if output matches format}, \\
0, & \text{if output doesn't match format}.
\end{cases}
\end{equation}

\paragraph{IoU reward in spatio-temporal
perception.} For the spatio-temporal
perception such as temporal grounding and object tracking, it requires the Video MLLM to output the time interval in the video that is associated with the content of a given textual query. Evidently, we can use the Intersection over Union (IoU) between the predicted interval by the model and the ground-truth interval as the reward function. This reward function effectively characterizes the accuracy of the interval predicted by the model.
\begin{equation}
    R_{\mathrm{IoU}} = \frac{|\mathcal{I}_{\mathrm{pred}} \cap \mathcal{I}_{\mathrm{gt}}|}{|\mathcal{I}_{\mathrm{pred}} \cup \mathcal{I}_{\mathrm{gt}}|},
\end{equation}
where $\mathcal{I}_{\mathrm{pred}}$ and $\mathcal{I}_{\mathrm{gt}}$ are the predicted and the ground truth of time intervals or detection boxes, respectively.

\paragraph{Accuracy reward in classification.} Discriminative tasks, such as multiple-choice video question answering and classification, aim to determine whether the model's prediction is consistent with the answer to the question. Therefore, we define:
\begin{equation}
    R_{\mathrm{accuacy}} = \begin{cases}
0, & \text{if } A_{\mathrm{pred}} \neq A_{\mathrm{gt}} \\
1, & \text{if } A_{\mathrm{pred}} = A_{\mathrm{gt}},
\end{cases}
\end{equation}
where $A_{\mathrm{pred}}$ and $A_{\mathrm{gt}}$ denote the predicted and the ground truth answers, respectively.

\paragraph{Recall reward in video captioning.} For tasks like video captioning with open-ended outputs, it is impossible to simply compare and determine the gap between the generated caption and the ground truth caption. Therefore, we use a LLM as a ``judge" to provide a reward score. In order to reduce the uncertainty of the evaluation criteria for the LLM, we first make the LLM decompose the ground truth and predicted captions into events list. Specifically, we utilize Qwen2.5-72B~\cite{qwen25} to extract the events in the description and judge whether the events in a ground truth description can be entailed by the description predicted by the model. We calculate the event recall score as the ratio of events in a ground truth description that are entailed by the predicted description, and set different rewards according to the event recall score:
\begin{equation}
    R_{\mathrm{recall}} = \mathrm{Recall_{event}}(C_{\mathrm{pred}}, C_{\mathrm{gt}}),
\end{equation}
where $C_{\mathrm{pred}}$ and $C_{\mathrm{gt}}$ represent the predicted and the ground truth captions, respectively.

By combining the above reward functions, we explored how to utilize GRPO to enhance the performance of Video MLLM in various tasks. The specific details can be found in the Section \ref{sec:Experiments}.

\subsection{Enhance Spatio-Temporal Perception of Video MLLM through GRPO}

\paragraph{The combination of reward functions.} We adopt different combinations of reward functions for training in different tasks. Specifically, for the temporal grounding and object tracking task, \(R_{\mathrm{st}}=R_{\mathrm{format}} + R_{\mathrm{IoU}}\). For the multi-choice QA and video quality assessment, \(R_{\mathrm{qa}}=R_{\mathrm{format}} + R_{\mathrm{accuacy}}\). For the multi-choice QA with glue (e.g. Grounding QA), \(R_{\mathrm{gqa}}=R_{\mathrm{format}} +R_{\mathrm{IoU}}+ R_{Acc}\). For the video caption, \(R_{\mathrm{cap}}=R_{\mathrm{format}} + R_{\mathrm{Caption}}\). 

\paragraph{VideoChat-R1-18k.} We collect diverse video corpus from existing public video datasets for reinforcement learning training to enhance the model's spatiotemporal perception ability. For the temporal grounding task, we use the training set of Charade - STA~\cite{charadesta} (5,338 samples) for training. For the object tracking task, training is conducted on the GoT - 10k~\cite{got} dataset, which has 9,335 samples. For the QA and grounding QA tasks, the validation set of NExTGQA ~\cite{nextgqa} (3,358 samples) is used for training. For video captioning, FIBER-1k~\cite{fiber1k} (1,000 samples) is adopted for training. For video quality assessment, we use the quality assessment task from VidTAB~\cite{videoeval} under the 100-shot setting, with 200 samples for training. Finally, for the training of VideoChat-R1, we perform joint training on three spatio-temporal perception-related tasks: temporal grounding, object tracking, and grounding QA. In total, 18,031 samples are used for training.

\begin{algorithm}[H]
\caption{VideoChat $\mathcal{M}$ with “Clue-Perception"}
\label{alg:re-perception}
\SetAlgoLined
\DontPrintSemicolon
\KwIn{
    $V_{\text{low}}$: Low-resolution/low-fps video \\
    $Q$: User question \\
    $\Delta_{\text{res}}$: Resolution boost factor \\
    $\Delta_{\text{fps}}$: Frame rate boost factor
}
\KwOut{$A^{(2)}$: Final refined answer}

\BlankLine
\textbf{Initial Answer Generation:} \\
\Indp
    $(A^{(1)}, \mathcal{C}) \gets \mathcal{M}(V_{\text{low}}, Q)$ 
\Indm

\BlankLine
\textbf{Clue Processing \& Upsampling:} \\
\Indp
    $\mathcal{T}_c \gets \text{ExtractTemporalClues}(\mathcal{C})$ \\
    $V_{\text{seg}} \gets \text{SelectSegments}(V_{\text{low}}, \mathcal{T}_c)$ \\
    $V_{\text{high}} \gets \text{Upsample}(V_{\text{seg}}, \Delta_{\text{res}}, \Delta_{\text{fps}})$ 
\Indm

\BlankLine
\textbf{Final Answer Generation:} \\
\Indp
    $A^{(2)} \gets \mathcal{M}(V_{\text{high}}, Q)$ 
\Indm

\BlankLine
\Return $A^{(2)}$ 
\end{algorithm}

\paragraph{Chat with Enhanced Spatio-Temporal Perception.}  After enhancing the spatiotemporal perception capability of MLLMs, we can construct a more reliable video chat system. Specifically, after the model answers a user's question, it can provide relatively accurate clues that support the answer to that question. We can further leverage these clues to improve the accuracy of the responses. Herein, we propose a simple \textbf{``Temporal Clue-driven Reasoning"} schema: after the model gives the first answer, we re-input the video segments corresponding to the obtained clues into the model at a higher resolution and frame rate, prompting it to answer again. By perceiving more details, the model can generate more accurate responses. Additionally, this operation is also well-suited for long video understanding under conditions of limited computing resources.

\section{Experiments}
\label{sec:Experiments}
\paragraph{Implementation details.} The main experiments are all conducted based on Qwen2.5-VL-7B~\cite{qwen25vl} (except for the video captioning, for which Qwen2-VL-7B~\cite{qwen2vl} is used). 

\paragraph{Benchmarks.}  We employ MVBench~\cite{li2024mvbench}, Perception Test~\cite{patraucean2023perception}, VideoMME~\cite{videomme} for evaluation of general video understanding. Given that the majority of videos in our training set are short-length, we only use the short subset of VideoMME in testing. For the temporal grounding task, we use the test set of Charade-STA~\cite{charadesta} for in-domain testing and the test set of ActivityNet-Grounding~\cite{anetgrounding} as out-domain test data. For the object tracking task, testing is done using the GoT-10k~\cite{got} dataset. For the QA and grounding QA tasks, the test set of NExTGQA ~\cite{nextgqa} is used for testing. And we use Dream-1k~\cite{dream1k} and VidTAB-QA~\cite{videoeval} for the video captioning and video quality access.

\subsection{Evaluation of VideoChat-R1}

\begin{table*}[!htbp]
    \centering
\begin{adjustbox}{width=\linewidth,center}
\renewcommand{\arraystretch}{1.1}
\setlength{\tabcolsep}{1.5mm}
\begin{tabular}{l|ccc|ccc|cc|cc|ccc}
\toprule 

\multirow{2}{*}{\textbf{Method}} & \multicolumn{3}{c}{\textbf{Charades-STA} }  & \multicolumn{3}{c}{\textbf{ActivityNet}} &  \multicolumn{2}{c}{\textbf{ NExTGQA }}  &  \multicolumn{2}{c}{\textbf{GoT }}    & {\textbf{VideoMME} }  & {\textbf{MVBench} }  & {\textbf{Peception Test} }   \\ 
  & mIoU & R@0.5 & R@0.7 & mIoU & R@0.5 & R@0.7 & mIoU & acc & Overlap & R@0.5  & Short-Avg & Avg &  Val \\

\midrule
\multicolumn{12}{l}{\darkGreen{\textit{Baseline}}}  \\
Qwen2.5-VL-7B &  29.0 & 24.2 & 11.1 & 21.1 & 15.8 & 7.5 & 15.4 & 59.5  & 12.6 & 1.1 & 71.3  & 66.9 & 69.1  \\

\midrule
\multicolumn{12}{l}{\darkGreen{\textit{SFT on specific tasks}}}  \\
+SFT w/ Charades-STA  &  46.3  & 45.0 & 25.3 & 20.6  & 16.7 & 7.9 & - & - & - & - & N/A* & N/A* &  N/A* \\
+SFT w/ GoT &  - & - & - & - & - & - & - & -  & 41.8 & 29.5 & 59.2  & 58.6 &  58.5 \\
+SFT w/ NExTGQA &  - & - & - & - & - & - & 28.2 & 64.8  & - & - & 60.1 & 59.2 & 60.7 \\
\midrule
\multicolumn{12}{l}{\darkGreen{\textit{GRPO on various tasks}}}  \\
\textbf{VideoChat-R1} &  \textbf{60.8} & \textbf{71.7} & \textbf{50.2} & \textbf{36.6} & \textbf{33.4} & \textbf{17.7} &\underline{32.4} & \textbf{70.6} & \textbf{43.8} & \textbf{38.2} & \underline{72.2} & \textbf{67.9}& \textbf{70.0}\\
\textbf{VideoChat-R1-thinking} &  \underline{59.9} & \underline{70.6} & \underline{47.2} & \underline{35.5} & \underline{33.3} & \underline{16.7} &\textbf{36.1} & \underline{69.2} & \underline{43.3} & \underline{33.9} & \textbf{74.2} & \underline{66.2}& \underline{69.6}\\

\bottomrule
\end{tabular}
\end{adjustbox}
\caption{\textbf{Results of VideoChat-R1 on various Video Benchmarks.} * indicates that the model has suffered from overfitting and is unable to answer the question properly. Since the number of input pixels is fixed during our evaluation, the baseline results are slightly lower than those reported in their origin paper~\cite{qwen25vl}.}
\label{tab:sota}
\vspace{-2mm}
\end{table*}

As shown in Table \ref{tab:sota}, after training with GRPO on spatio-temporal perception datasets, both VideoChat-R1 and VideoChat-R1-thinking significantly outperform the performance of Qwen2.5-VL and that of models fine-tuned through SFT for a single specific task across various spatiotemporal perception benchmarks and the general understanding benchmark VideoMME. This validates the effectiveness of our approach, which leverages multiple spatiotemporal perception datasets and RFT for enhancing spatiotemporal perception.

Meanwhile, we observe that for spatio-temporal perception tasks, engaging in thinking processes does not necessarily lead to performance gains. However, for tasks such as QA and VideoMME, which may require complex reasoning, conducting inferences during testing can result in notable performance improvements.

\subsection{Ablation Studies and Discussions}

\paragraph{Muti-task Co-training.} As shown in Table \ref{tab:cotraining}, we found that mixed training of different spatiotemporal perception tasks using GRPO can yield a synergistic improvement effect. Training with the multiple tasks achieves nearly the best results across all benchmarks. This reveals the potential of GRPO for larger-scale and multi-task collaborative training in the future.

\begin{table*}[!htbp]
    \centering
\begin{adjustbox}{width=\linewidth,center}
\renewcommand{\arraystretch}{1.1}
\setlength{\tabcolsep}{1.5mm}
\begin{tabular}{l|ccc|ccc|cc|cc|c}
\toprule 
\multirow{2}{*}{\textbf{Method}} & \multicolumn{3}{c}{\textbf{Charades-STA} }  & \multicolumn{3}{c}{\textbf{ANet}} &  \multicolumn{2}{c}{\textbf{NExTGQA}}  &  \multicolumn{2}{c}{\textbf{GoT }}    & {\textbf{VideoMME} }  \\ 
  & mIoU & R@0.5 & R@0.7 & mIoU & R@0.5 & R@0.7 & mIoU & acc & Overlap & R@0.5  & Short-Avg   \\

\midrule

Qwen2.5-VL-7B &  29.0 & 24.2 & 11.1 & 21.1 & 15.8 & 7.5 & 15.4 & 59.5  & 12.6 & 1.1 & 71.3   \\

+GRPO w/ STA &  59.3 & \underline{70.4} & 46.0 & 30.7 & 27.5 & 12.9 & 31.4  & 61.2 & 27.8 & 12.9 & \textbf{72.6}  \\

+GRPO w/GQA & 36.0  & 33.5 & 15.5 & 24.9& 20.6& 10.7& 35.1 & \underline{68.7} & 36.1  & 26.7 & 72.0   \\

+GRPO w/ GoT & 28.7 & 25.1 & 9.6 & 20.1 & 16.2  & 6.8 & 15.6  & 60.5 & \underline{42.5} & \underline{30.6} &  71.4 \\

+GRPO w/ STA-GQA &  \underline{59.8} & 69.7  & \underline{47.0} & \underline{33.7} & \underline{31.0} & \underline{16.0} & \textbf{35.7} & 67.7 & 36.5 & 28.9 & \underline{72.2} \\

+GRPO w/ STA-GQA-GoT &  \textbf{60.8} & \textbf{71.7} & \textbf{50.2} & \textbf{36.6} & \textbf{33.4} & \textbf{17.7} & 32.4 & \textbf{70.6} & \textbf{43.8} & \textbf{38.2} &  \underline{72.2} \\

\bottomrule
\end{tabular}
\end{adjustbox}
\caption{\textbf{Ablation results of Cotraining on Spatio-Temporal Tasks.}}
\label{tab:cotraining}
\end{table*}

\begin{table*}[!htbp]
    \centering
\begin{adjustbox}{width=\linewidth,center}
\renewcommand{\arraystretch}{1.1}
\setlength{\tabcolsep}{1.5mm}
\begin{tabular}{l|c|cc|cc|llll|llll|l}
\toprule 

\multirow{2}{*}{\textbf{Method}} & \multirow{2}{*}{\textbf{Epochs}} & \multicolumn{2}{c}{\centering \textbf{Training Prompt}} & \multicolumn{2}{c}{\centering \textbf{Test Prompt}}& \multicolumn{4}{c}{\textbf{Charades-STA (in domain)}}  &\multicolumn{4}{c}{\textbf{ActivityNet (out domain)} }   & {\textbf{VideoMME} } \\ 
 & & \centering{Think} & \centering{Answer} & \centering{Think} & \centering{Answer}  & mIoU & R@0.3 & R@0.5 & R@0.7 & mIoU & R@0.3 & R@0.5 & R@0.7 & Short-Avg  \\

\midrule
\multicolumn{8}{l}{\darkGreen{\textit{Vision Experts}}}  \\

FlashVTG~\cite{cao2024flashvtg}  & - &  -& - & - & -& - &- & 70.3 & 49.9 & -& -& -& - & -\\
InternVideo2-6B~\cite{wang2024internvideo2}   &  -& - &-  & - &- & -& -& 70.0 & 49.0 &-&-&-& -& -\\
SG-DETR~\cite{gordeev2024saliency}  &  &  &  &  & & -&- & 71.1 & 52.8 &-&-&-&- & -\\
\midrule
\multicolumn{8}{l}{\darkGreen{\textit{MLLMs}}}  \\

Qwen2.5-VL-7B & - & - & - &  & \checkmark & 29.0 & 44.7 & 24.2 & 11.1 & 21.1& 28.3 & 15.8 & 7.4 &  71.3  \\
    (baseline) & - & - & - & \checkmark & \checkmark & 28.1 & 41.8 & 23.4 &11.1 & 17.7 &22.7 & 13.4 &7.7 & 71.3  \\

\midrule
+ SFT & 1 &  & \checkmark &  & \checkmark & 46.3 & 63.9 & 45.0 & 25.3 &20.6 & 30.2  & 16.7 & 7.9 & N/A*\textcolor{teal}{(-71.3)} \\
 & 3 &  & \checkmark &  & \checkmark & 34.6\textcolor{red}{(+6.5)} & 51.7 & 36.3 & 20.6 & 17.3\textcolor{teal}{(-3.8)} & 26.1 & 10.0 & 3.9 & N/A*\textcolor{teal}{(-71.3)} \\
\midrule

+ GRPO & 1 &  & \checkmark &  & \checkmark & 58.7 & 80.9 & 67.7 & 45.4 & 31.9 & 46.3 & 28.8 & 14.1 & 72.6 \\

& 1 & \checkmark & \checkmark &  \checkmark  & \checkmark & 59.3\textcolor{red}{(+31.2)}  & 81.7 & 70.4 & 46.0 & 30.7\textcolor{red}{(+13.0)} & 45.0 & 27.5& 12.9 & 73.6\textcolor{red}{(+2.3)} \\

& 3 & \checkmark & \checkmark & \checkmark & \checkmark & 61.3\textcolor{red}{(+33.2)} & 83.1 & 72.8 & 51.5 & 34.3\textcolor{red}{(+16.6)} & 50.4 & 32.2 & 16.2 & 70.9\textcolor{teal}{(-0.4)} \\

\bottomrule
\end{tabular}
\end{adjustbox}
\caption{\textbf{Ablation results of Temporal Grounding Task.} * indicates that the model has suffered from overfitting and is unable to answer the question properly.}
\label{tab:grounding}
\vspace{-2mm}
\end{table*}

\begin{table}[!htbp]
    \centering

\renewcommand{\arraystretch}{1.1}
\setlength{\tabcolsep}{1.5mm}
\centering
\begin{adjustbox}{width=0.6\linewidth,center}
\begin{tabular}{l|lll|lll}
\toprule 

\multirow{2}{*}{\textbf{Method}} & \multicolumn{3}{c}{\textbf{GoT} } & {\textbf{VideoMME} }  \\ 
  & Average overlap & R@0.5 & R@0.7  & Short-Avg   \\

\midrule

Qwen2.5-VL-7B &  12.6 & 1.1 & 0 & 71.3 \\
+SFT &  41.8 & 29.5 & 3.9  & 59.2 \\
+GRPO &  42.5\textcolor{red}{(+29.9)} & 30.6\textcolor{red}{(+29.5)} & 3.9\textcolor{red}{(+3.9)} & 71.4\textcolor{red}{(+0.1)} \\

\bottomrule
\end{tabular}

\end{adjustbox}
\caption{\textbf{Ablation results of Object Tracking.} We use 8 frames as input for training and evaluation.}
        \label{tab:tracking}

\label{tab:main}
\vspace{-2mm}
\end{table}
\paragraph{Temporal Grounding and Object tracking.}

As shown in Table \ref{tab:grounding} and Table \ref{tab:tracking}, fine-tuning Qwen2.5-VL using GRPO significantly improves the performance of temporal grounding and object tracking tasks. Additionally, it slightly enhances the performance on the general understanding benchmark VideoMME. Even when training for more epochs, GRPO is less prone to overfitting compared to SFT. Instead, it can continuously improve the performance of temporal grounding, eventually surpassing the performance of previous expert models. Moreover, stimulating the model's thinking ability provides some benefits for both temporal grounding and VideoMME tasks.

\begin{table*}[!htbp]
    \centering
\begin{adjustbox}{width=0.8\linewidth,center}
\renewcommand{\arraystretch}{1.1}
\setlength{\tabcolsep}{1.5mm}
\begin{tabular}{l|ccc|ccc|ll|l}
\toprule 
\multirow{2}{*}{\textbf{Method}} & \multicolumn{3}{c}{\multirow{1}{*}{\centering \textbf{Training Prompt}}} & \multicolumn{3}{c}{\multirow{1}{*}{\centering \textbf{Test Prompt}}}& \multicolumn{2}{c}{\textbf{ NExTGQA} }   & {\textbf{VideoMME} } \\ 
& Think & Answer & Glue & Think & Answer & Glue  & \centering{mIoU} & \centering{acc} & Short-Avg  \\

\midrule
\multicolumn{10}{l}{\darkGreen{\textit{Direct Output}}}  \\

Qwen2.5-VL-7B &  &  & & &\checkmark & & - & 41.7 & 71.3   \\
    (baseline)   &  &  & & &\checkmark & \checkmark & 15.4 & 59.5 &   - \\

\midrule
+ SFT & & \checkmark &  & & \checkmark & & - & 65.1 & 60.2  \\
          & & \checkmark & \checkmark &  & \checkmark & \checkmark & 28.2\textcolor{red}{(+12.8)} & 64.8\textcolor{red}{(+5.3)} & 60.1\textcolor{teal}{(-11.2)}  \\
\midrule
+ GRPO & & \checkmark &  & & \checkmark & & - &  70.1 & 71.7  \\
          & & \checkmark &  & & \checkmark & \checkmark & 16.2 & 70.2 & 71.7  \\

& & \checkmark & \checkmark & & \checkmark &  \checkmark & 35.1\textcolor{red}{(+19.7)} & 68.7\textcolor{red}{(+9.2)} & 72.0\textcolor{red}{(+0.7)}  \\
\midrule
\multicolumn{10}{l}{\darkGreen{\textit{Chain-of-thought Output}}}  \\
Qwen2.5-VL-7B & &  & &  \checkmark &  \checkmark & &  - & 47.7 & 73.0  \\
            & &  & &  \checkmark &  \checkmark &  \checkmark &  20.2 & 53.3 &  72.2 \\
\midrule
+ GRPO & \checkmark & \checkmark &  & \checkmark & \checkmark & & - & 68.8 & 74.7  \\
            & \checkmark & \checkmark & \checkmark & \checkmark & \checkmark & \checkmark & 32.9\textcolor{red}{(+12.7)} &  66.9\textcolor{red}{(+13.6)}  & 75.3\textcolor{red}{(+3.1)}  \\
\bottomrule
\end{tabular}
\end{adjustbox}
\caption{\textbf{Ablation results of Multi-Choice Video QA.}}
\label{tab:gqa}
\vspace{-2mm}
\end{table*}
\paragraph{Video Question Answer.} As shown in Table \ref{tab:gqa}, for the video question answering task, we selected the multi-choice QA task, which is easy to evaluate, for our experiments. Additionally, we explored the grounding QA task. In this task, when answering questions, the model is required to simultaneously provide the temporal cues on which its answers are based. Using merely a little over three thousand training data samples, we found that GRPO demonstrated remarkable fine-tuning capabilities. Not only did it lead to a substantial improvement in the performance of the  NExTGQA  task, but it also brought about a noticeable enhancement in the VideoMME task. We noticed that, unlike the previous strongly spatiotemporal perception tasks such as temporal grounding, thinking played a significant role in the QA task. Meanwhile, the glue signals also provided some assistance for relatively complex video understanding tasks like VideoMME.

\begin{table}[!htbp]
    \centering
\begin{adjustbox}{width=0.6\linewidth,center}
\renewcommand{\arraystretch}{1.1}
\setlength{\tabcolsep}{1.5mm}
\begin{tabular}{l|ccc|c}
\toprule 
\multirow{2}{*}{\textbf{Method}} & \multicolumn{3}{c}{\textbf{Dream-1k} }& \multicolumn{1}{c}{\textbf{VidTAB-QA} } \\ 
 &F1 & Precision &  Recall  & Accuracy   \\

\midrule

Baseline & 30.6 & 33.8  &  27.9  & 70.7   \\
\midrule
+ SFT & 31.4 & 32.6  & 30.2  &  71.7  \\
\midrule
+ GRPO & 38.2\textcolor{red}{(+7.6)} & 45.4\textcolor{red}{(+11.6)} & 33.1\textcolor{red}{(+5.2)}  &  72.6\textcolor{red}{(+1.9)}  \\
\bottomrule
\end{tabular}
\end{adjustbox}
\caption{\textbf{Results of Video Caption and Video Quality Access.}}
\label{tab:vidtabqa}
\vspace{-2mm}
\end{table}
\begin{table}[!htbp]
    \centering
\begin{adjustbox}{width=0.6\linewidth,center}
\renewcommand{\arraystretch}{1.1}
\setlength{\tabcolsep}{1.5mm}
\begin{tabular}{@{}c|c|ccc@{}}
\toprule
\multirow{2}{*}{\textbf{Method}} & \multirow{2}{*}{\begin{tabular}[c]{@{}c@{}}LLM\\ Judge\end{tabular}} & \multicolumn{3}{c}{\textbf{Dream-1k}}                                                          \\ \cmidrule(l){3-5} 
                                 &                                                                      & F1                            & Precision                      & Recall                        \\ \midrule
Baseline                         & -                                                                    & 30.6                          & 33.8                           & 27.9                          \\ \midrule
\multirow{2}{*}{+ GRPO}          & GPT-3.5-turbo-0125                                                   & 37.9\textcolor{red}{(+7.3)} & 44.4\textcolor{red}{(+10.6)} & 33\textcolor{red}{(+5.1)}   \\
                                 & Qwen2.5-72B                                                          & 38.2\textcolor{red}{(+7.6)} & 45.4\textcolor{red}{(+11.6)} & 33.1\textcolor{red}{(+5.2)} \\ \bottomrule
\end{tabular}
\end{adjustbox}
\caption{\textbf{Ablation of Video Caption Task.}}
\label{tab:videcaption_judge}
\vspace{-2mm}
\end{table}

\paragraph{Video Caption and Video Quality Assessment.} For the Video Caption and Video Quality Assessment tasks, we found that GRPO still demonstrated its advantages over SFT, As shown in Table \ref{tab:vidtabqa}. The significant metric improvements on these two benchmarks demonstrate the effectiveness of our approach.

\paragraph{Ablation of Reward Evaluators}
To assess the impact of different large language models (LLMs) as reward evaluators, we conducted parallel experiments using GPT-3.5-turbo-0125 and Qwen2.5-72B as distinct judges (Table \ref{tab:videcaption_judge}). Models trained under both evaluators achieved nearly identical performance, demonstrating consistent caption reward generation across LLMs. We attribute this consistency to GRPO's fundamental mechanism: it relies on relative differential scoring within response groups rather than absolute reward values. This confirms that $R_{recall}$ produces discriminative reward signals for predicted captions independent of the choice of LLM judge, validating both the efficacy of our reward design and the stability of its signaling mechanism. Crucially, when guided by these reliable reward signals, our approach delivers substantial performance gains in description tasks using only limited high-quality data, demonstrating remarkable data-efficiency and significant optimization potential.

\begin{table}[!htbp]
    \centering

\renewcommand{\arraystretch}{1.1}
\setlength{\tabcolsep}{1.5mm}
\centering
\begin{adjustbox}{width=0.75\linewidth,center}
\begin{tabular}{l|l|ll}
\toprule 

\textbf{Model} & \textbf{Clue}  & \textbf{VideoMME}  & \textbf{LongVideoBench}   \\ 
 Avg. Duration & \textbf{Perception} & 1010s & 473s  \\

\midrule

Qwen2.5-VL-7B &   &  64.4  & 56.0 \\
 & \checkmark   & 63.3\textcolor{teal}{(-1.1)}    & 55.2\textcolor{teal}{(-0.8)}  \\
 \hline
\textbf{VideoChat-R1-thinking} &   & 62.1  & 51.9 \\
 &  \checkmark & 63.6\textcolor{red}{(+1.5)}  & 58.2\textcolor{red}{(+6.3)} \\

\bottomrule
\end{tabular}

\end{adjustbox}
\caption{\textbf{Ablation results of “Clue-Perception".} It should be noted that due to our adoption of a lower number of input pixels, the absolute performance is not entirely consistent with that reported for Qwen2.5-VL.}
        \label{tab:reperception}

\label{tab:main_grounding}
\vspace{-2mm}
\end{table}

\paragraph{Ablation of ``Clue-Perception"} As shown in Table. \ref{tab:reperception}, we compared the performance changes of the model with and without perception enhancement when applying the "Clue-Perception" strategy on two representative long video benchmarks~\cite{videomme,longvideobench}. It is noteworthy that without the use of "Clue-Perception", VideoChat-R1 showed no significant performance improvement over Qwen2.5-VL-7B in long video tasks, which can be attributed to the fact that our training dataset consists entirely of short videos under 1 minute. However, after the application of the "Clue-Perception" operation, VideoChat-R1 demonstrated a significant performance enhancement, indicating that the clues it provides are more accurate and thus revealing the potential of clue-perception in long video understanding. In contrast, due to its insufficient spatiotemporal perception capability, Qwen2.5-VL-7B even exhibited a performance decline after the implementation of the "Clue-Perception" operation.

\paragraph{GRPO vs. SFT.} It can be observed that across various types of tasks, GRPO outperforms SFT. Whether it is in terms of the performance on in-domain tasks, out-domain tasks, or the preservation of the original general performance, our experimental results demonstrate that GRPO is a promising fine-tuning approach. We will leave the large-scale comparison for future research.

\paragraph{Chain-of-thought vs. Direct Output.} Based on the video tasks and experiments we have explored, which focus on spatiotemporal perception, the output of the chain of thought has not demonstrated obvious advantages. In some cases, it is even inferior to the direct output. We believe that how to define appropriate video reasoning tasks and evaluation methods remains to be explored. The existing training data is insufficient to activate the model to output truly effective video reasoning chains.

\begin{figure*}[!h]
	\centering
	\includegraphics[width=0.88\textwidth]{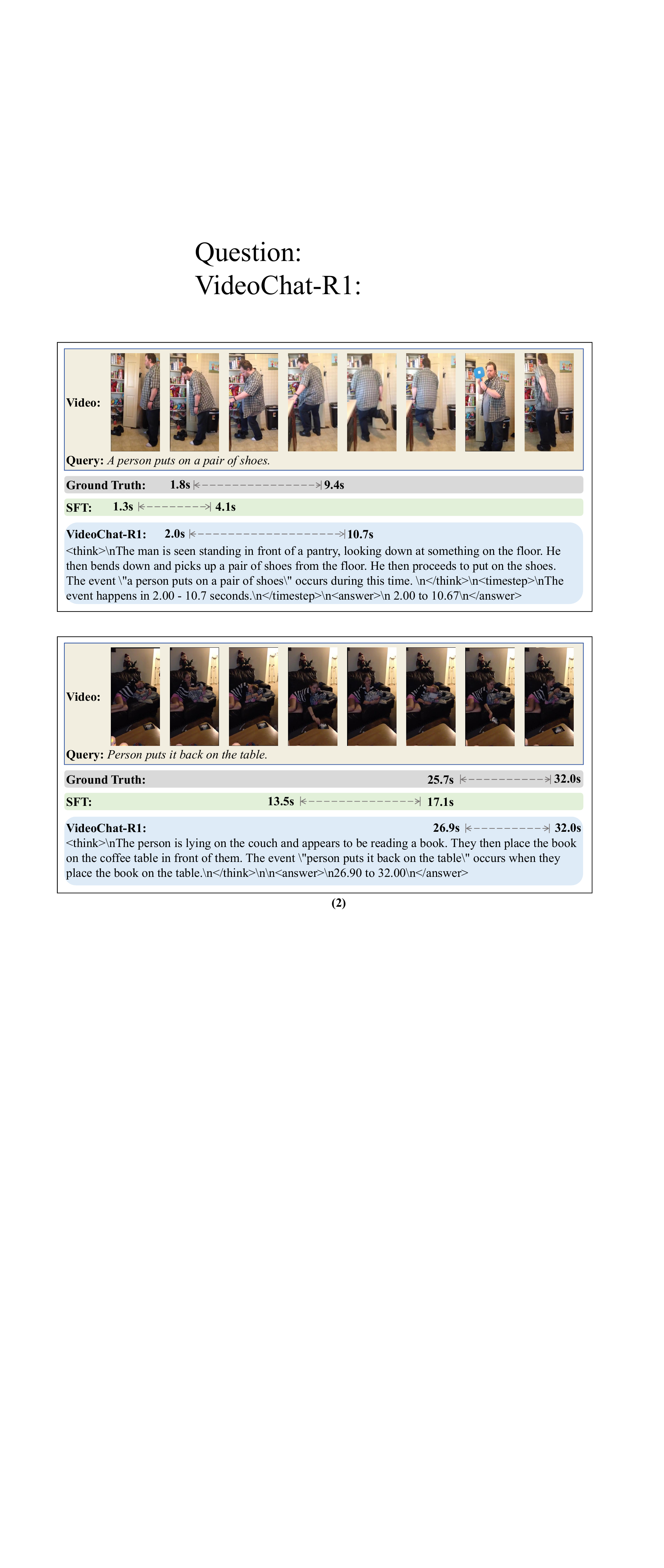} 
	\caption{
    \textbf{Examples on temporal grounding task.} VideoChat-R1 gives a more accurate time interval after thinking.
    }
	\label{fig2:example}
\end{figure*}

\begin{figure*}[!h]
	\centering
	\includegraphics[width=0.88\textwidth]{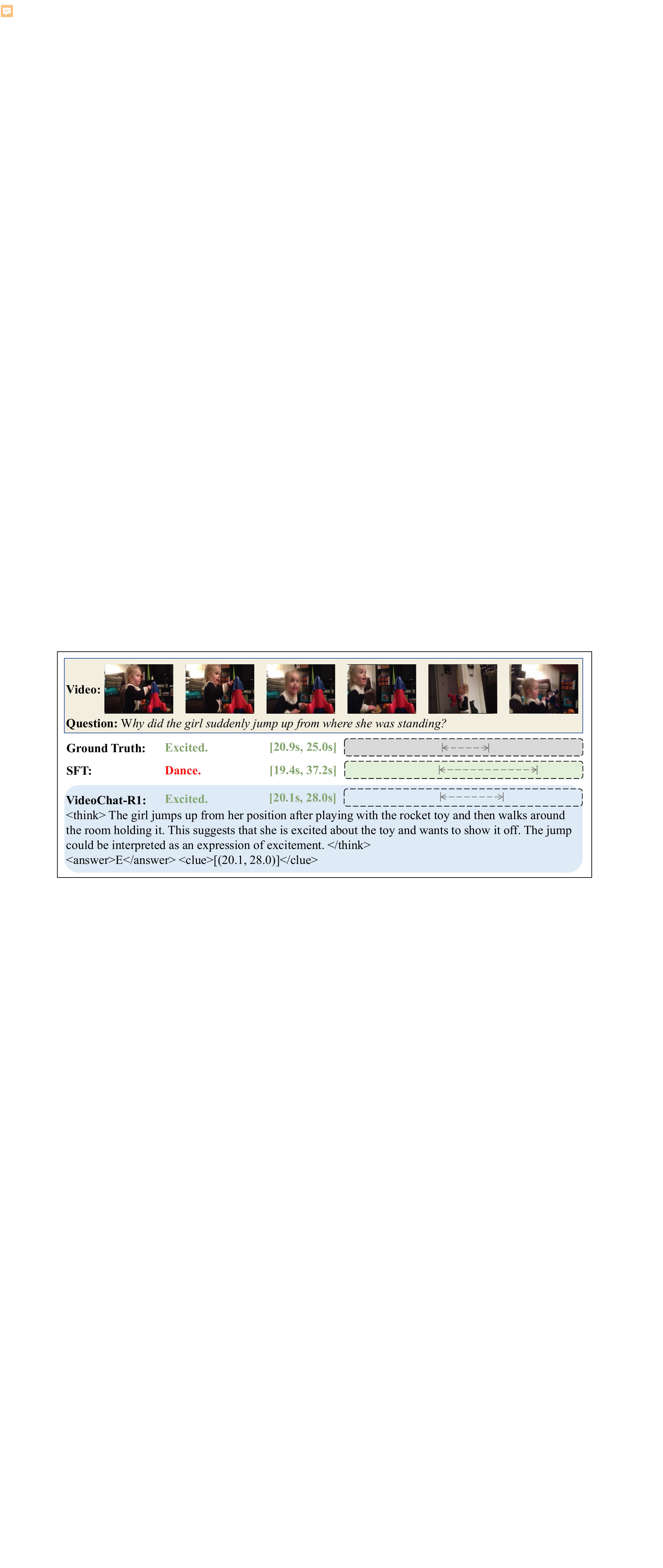} 
	\caption{
   \textbf{Examples on Video QA task.} It can be seen that VideoChat-R1 can not only answer questions correctly but also provide relatively accurate reference time periods (clue).}
	\label{fig2:grounding_case}
\end{figure*}


\subsection{Qualitative Results}

As shown in Figure \ref{fig2:example} and \ref{fig2:grounding_case}, we provide visualizations of VideoChat-R1's outputs for the temporal grounding and video QA tasks. We observe that due to the enhanced spatiotemporal perception capability, VideoChat-R1 can more accurately locate video segments relevant to the question and conduct reasonable reasoning to answer the question. However, compared with the complex chains of thought involved in solving mathematical problems or coding tasks in text and image domains, the chains of thought generated in our current video tasks are relatively simple. We will explore more complex video reasoning tasks in future research.

\section{Conclusions}
\label{sec:Conclusions}

In this work, we systematically investigate the role of reinforcement fine-tuning (RFT) with Group Relative Policy Optimization (GRPO) in enhancing video-centric multimodal large language models (MLLMs). Our experiments demonstrate that RFT is a highly data-efficient paradigm for task-specific improvements, enabling VideoChat-R1—a model trained with limited samples via multi-task RFT—to achieve state-of-the-art performance on spatio-temporal perception tasks while preserving general chat capabilities and exhibiting emergent spatiotemporal reasoning. We believe our work can present relevant insights for future research efforts in reinforcement learning of video MLLMs.

\clearpage
{
\small
\bibliographystyle{plainnat}
\bibliography{reference}
}

\end{document}